\documentclass[letterpaper, 10 pt, journal, twoside]{IEEEtran} 

\IEEEoverridecommandlockouts                              

\usepackage{amsmath} 
\usepackage{amssymb}  
\usepackage{graphicx}
\usepackage{subfig}
\usepackage{xcolor}
\usepackage{flushend}
\usepackage{perl_acronyms}
\usepackage[
    style=ieee,
    doi=false,
    isbn=false,
    url=false,
    eprint=false,
    backend=bibtex,
    natbib=true
    ]{biblatex}
\title{Occlusion-Aware Risk Assessment for \\ Autonomous Driving in Urban Environments}

\author{Ming-Yuan Yu$^{1}$, Ram Vasudevan$^{2}$, and Matthew Johnson-Roberson$^{3}$

\thanks{This work was supported by a grant from Ford Motor Company via the Ford-UM Alliance under Award N022884. (Corresponding author: Ming-Yuan Yu.)} 
  
\thanks{$^{1}$M.-Y Yu is with Robotics Institute, University of Michigan, Ann Arbor, MI 48109, USA
        {\tt\footnotesize myyu@umich.edu}}%
\thanks{$^{2} $R. Vasudevan is with the Department of Mechanical Engineering, the University of Michigan, Ann Arbor, MI 48109, USA
        {\tt\footnotesize ramv@umich.edu}}%
\thanks{$^{3} $M. Johnson-Roberson is with the Department of Naval Architecture and Marine Engineering at the University of Michigan, Ann Arbor, MI 48109, USA
        {\tt\footnotesize mattjr@umich.edu}}%
\thanks{Digital Object Identifier (DOI): 10.1109/LRA.2019.2900453.}
}

\begin{document}

\maketitle


\begin{abstract}

Navigating safely in urban environments remains a challenging problem for autonomous vehicles. 
Occlusion and limited sensor range can pose significant challenges to safely navigate among pedestrians and other vehicles in the environment. 
Enabling vehicles to quantify the risk posed by unseen regions allows them to anticipate future possibilities, resulting in increased safety and ride comfort.
This paper proposes an algorithm that takes advantage of the known road layouts to forecast, quantify, and aggregate risk associated with occlusions and limited sensor range. 
This allows us to make predictions of risk induced by unobserved vehicles even in heavily occluded urban environments. The risk can then be used either by a low-level planning algorithm to generate better trajectories, or by a high-level one to plan a better route.
The proposed algorithm is evaluated on intersection layouts from real-world map data with up to five other vehicles in the scene, and verified to reduce collision rates by $4.8\times$ comparing to a baseline method while improving driving comfort.

\end{abstract}



\begin{figure}[!ht]
  \centering
  \includegraphics[width=0.75\linewidth,trim={4mm 4mm 4mm 13mm},clip]{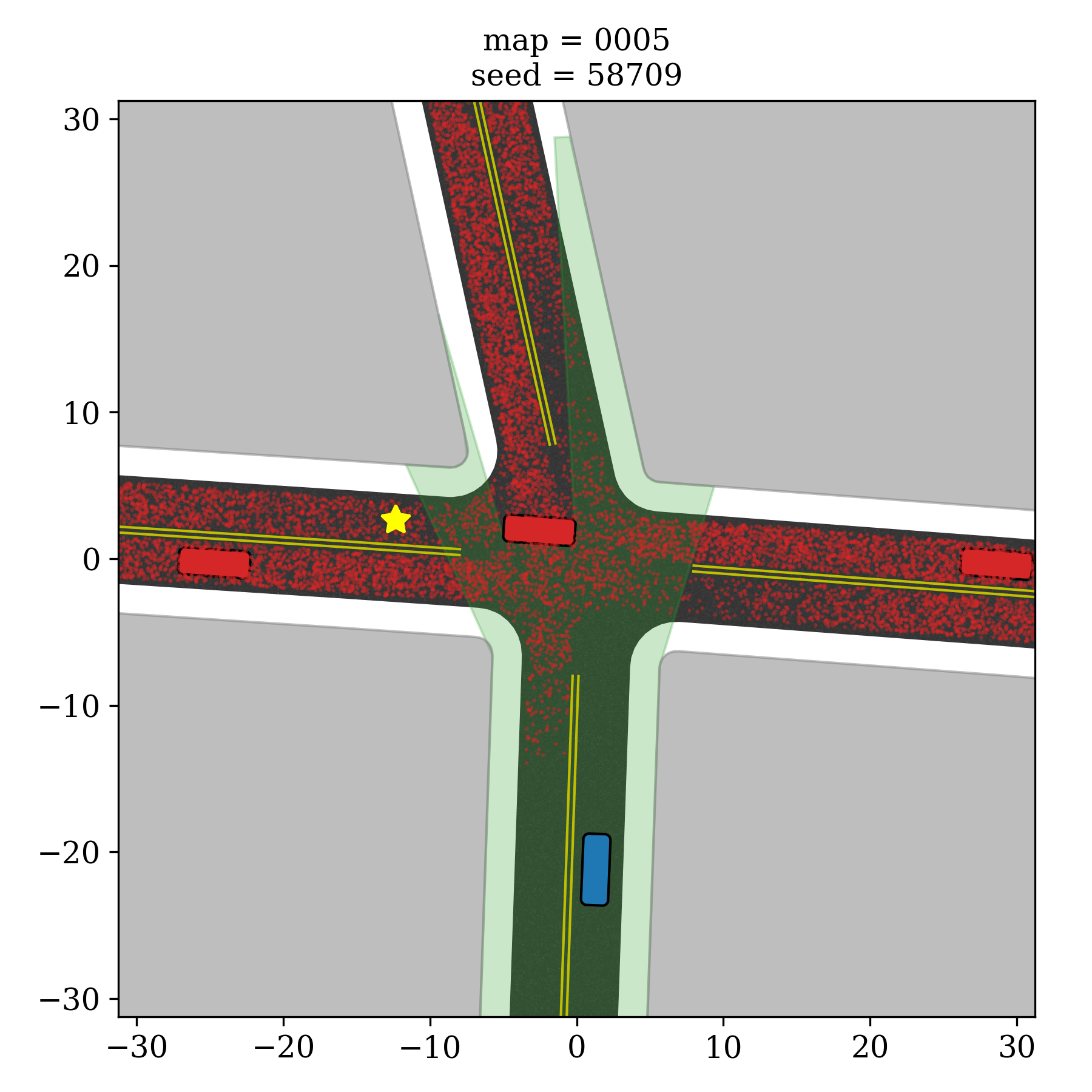}
  \caption{
  Ego vehicle (blue box) intends to perform an unprotected left turn to the goal (yellow star) at an intersection. The irregular shape of the observable polygon (green shaded region) is caused by 1) limited sensor range and 2) occlusions from other vehicles (red boxes) and buildings (gray regions.) Our algorithm quantifies the distribution of risk (red particles) posed by other vehicles including the ones which are outside of the observable polygon. This is possible under the assumption that we know the geometry of road layout and the nominal (or worst case) speed of other vehicles at this particular intersection. Both axes are in meters.}
  \label{fig:scene-ult}
\end{figure}

\section{INTRODUCTION}

\IEEEPARstart{A}{dvancements} in sensing technology and algorithmic improvements bring the reality of everyday autonomous driving closer to fruition. 
LIght Detection and Ranging (LIDAR) \acused{LIDAR} sensors enable the construction of 3D maps \cite{kammel2008lidar} and can see tens or hundreds of meters away, even at night \cite{sudhakar2017image}. 
High definition cameras capture images that can be used for tasks such as semantic segmentation  \cite{Shelhamer2017fully} and object detection \cite{liu2015single}.
Many tasks can be performed at levels surpassing that of humans thanks to recent developments in deep neural networks \cite{kaiming2015delving}. 

However, all sensors still have limited sensing capabilities. 
\acp{LIDAR} and cameras, for instance, have difficulty identifying objects beyond a certain distance due to finite range, sensitivity, and angular resolution. 
In addition, both of these sensors can not see through opaque objects which could results in large unobserved regions. 
An illustration of such a scenario is shown in Fig. \ref{fig:scene-ult}.

One of the reasons why human drivers can safely navigate even under occlusions is that they augment their sensing capabilities by leveraging semantic and geometrical information of the environment, and anticipate the need to slow down due to the potential risk of collision that arises due to occlusions \cite{orzechowski2018tackling, lee2017collision}. 
In addition, earlier braking would reduce the maximum deceleration which consequently leads to greater ride comfort. 

To provide an example from the real-world imagine the following: pulling up to a left turn next to a tall tree or building, similar to the scenario shown in Fig. \ref{fig:scene-ult}.
Typically a driver leans forward and pulls the car slightly ahead to see into oncoming traffic before completing the turn. 
In the driver's mind, they have a map of the unseen spaces and know that a car could emerge from beyond the current line of sight.
As a result, they proceed cautiously to try to improve their visibility and do not turn until they can confirm a sufficient gap in the traffic. 

This paper presents an algorithm that encodes this form of human driving by quantifying the risk caused by limited sensing capabilities and geometric occlusion. 
The proposed algorithm can be used to make autonomous vehicles navigate safely with improved ride comfort in urban environments, and it is agnostic to how the vehicle makes decisions.

The paper is organized as follows:
Section \ref{sec:relate-work} reviews related work in the field of risk assessment and planning under occlusion. Section \ref{sec:method} describes how our algorithm leverages the known road layout to quantify risk in the environment, and demonstrates how it can be easily integrated with a simple planning algorithm. Section \ref{sec:evaluation} introduces two baseline methods and our evaluation methodologies. Section \ref{sec:results} evaluates the proposed occlusion-aware method, and shows that statistically our algorithm performs significantly better in terms of collision rate and ride comfort on both synthetic and real-world intersections. Section \ref{sec:conclusions} concludes and discusses future directions of this work.

\begin{figure*}[t]
  \centering
  \includegraphics[width=\linewidth]{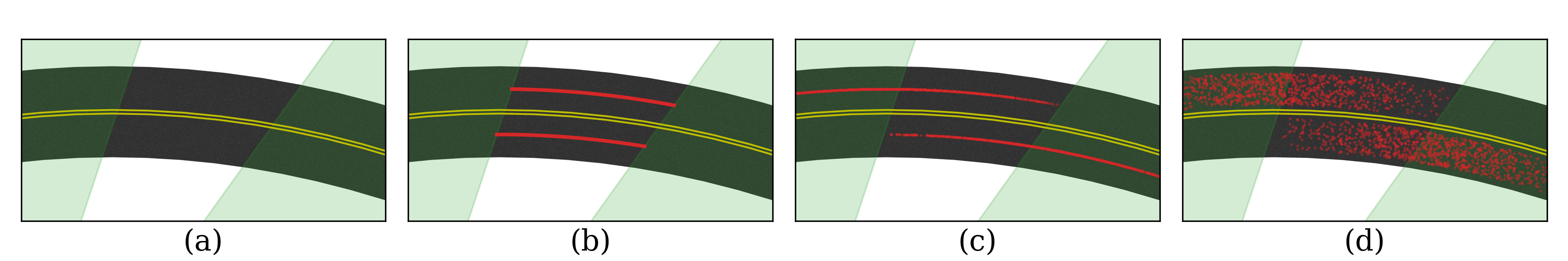}
  \caption{Illustration of our algorithm on a (a) partially observed road, where the green shaded regions are within the sensor's field-of-view. (b) Firstly, assuming that the map and the ego vehicle's location are known, the centerlines of the unobserved road segments (red) are extracted. (c) Secondly, we sample particles $\left\{(s^{[i]}, v^{[i]})\right\}_{i=1}^{N_k}$ along the extracted splines $c_k,k\in\{1,2\}$, where the location $s$ and speed $v$ are drawn from uniform distributions, and propagate each particle forward in time assuming constant speed. (d) Finally, a random offset perpendicular to each centerline is added to each particle to incorporate the non-zero size of potentially unobserved vehicles.}
  \label{fig:steps}
\end{figure*}

\section{RELATED WORK}
\label{sec:relate-work}

Most of the previous work on motion prediction and risk assessment can be categorized into one of two following categories. 
The first category quantifies the risk as the probability of having a collision with another vehicle or pedestrian. 
The second category assesses risk as the degree of deviation from a nominal set of behaviors (e.g. veering from the lane rapidly). 
A well-organized survey is given in \cite{lefevre2014survey}. 
This paper addresses the first category of problems with a specific focus on collisions caused by occluded objects.

Prior work has addressed the issue of occlusion from a tracking perspective. \citet{wyffels2015negative} keep track of obstacles in occluded areas by utilizing negative information under the assumption that undetected objects are not likely to appear in visible space.  \citet{yu2012shadow} maintain the tracks of targets that move outside the field-of-view by formulating the problem as a pursuit evasion game. \citet{galceran2015augmented} augment states of a standard tracker to estimate occluded states for other agents and provide more robust data association when the occluded agents reappear in the scene. \citet{ondruska2016deep} and \citet{dequaire2018deep} utilize recurrent neural networks trained in an unsupervised manner and are able to track occluded object from only raw, occluded sensor data. Although these models keep tracks of missing targets that enter occluded regions, they all need at least one detection to start tracking. They do not explicitly handle risks caused by potential incoming traffic which is occluded or outside the sensor horizon and thus never detected in the first place.

Partially Observable Markov Decision Process (POMDP) is a common approach to tackle decision making problems under uncertainty and consequently can implicitly handle probabilistic occlusion. \citet{Brechtel2014probabilistic} use Monte Carlo Value Iteration and \citet{Brechtel13solvingcontinuous} show it is possible to optimize a continuous POMDP model. \citet{Bouton2018scalable} approximate the global solution by solving a POMDP for each agent independently through utility fusion~\cite{Russell2003QRL}. The reduction in state space required to make these approaches tractable limits their applicability, particularly for real-time high speed driving. The algorithm we present in this paper differs in both goal and implementation. We focus on quantifying risk in the environment instead of the risk associated with the actions of the ego vehicle. Our algorithm is agnostic to planning and so could be coupled with a POMDP or any other planning or decision making algorithms.

Most closely related to the approach presented here are two risk quantification approaches~\cite{orzechowski2018tackling,lee2017collision}. \citet{orzechowski2018tackling} over-approximate all possible states of the incoming traffic by considering the leading edges of the visible polygon. Although safety is guaranteed, the resulting over-approximated polygons are not probabilistic, whereas our approach captures the full distribution of risk. \citet{lee2017collision} perform probabilistic risk assessment by utilizing prebuilt high definition maps. While the results in \cite{lee2017collision} look promising, they do not show how their risk assessment could be used for planning to achieve safer driving. Furthermore, both \cite{orzechowski2018tackling} and \cite{lee2017collision} show very limited results with only a single additional vehicle in the scene and it is unclear how these approaches perform in crowded scenes such as urban intersections. We focus on realistic intersections derived from real map data and occupied with many vehicles. 

Our approach presents several novel contributions: 1) we present an algorithm which performs probabilistic risk assessment of both observed and unobserved regions at urban intersections; 2) the approach is control algorithm agnostic and can be integrated with any deterministic or probabilistic planning approach; 3) we derive risk assessment from large-scale map data and extensively evaluate our approach with up to five other vehicles in the scene, and show significant reduction in collision rate and increase in ride comfort.

\section{METHOD}
\label{sec:method}

We first describe our method in probabilistic risk assessment in Section \ref{sec:ra-cartesian}. We generate a distribution over the Cartesian space. In Section \ref{sec:ra-action} we show how to integrate the risk to a simple optimization-based planning algorithm and describe the primary cost function.

\subsection{Risk Assessment Over Cartesian Space}
\label{sec:ra-cartesian}

High Definition (HD) maps are used commonly in autonomous driving~\cite{lee2017collision}. These maps have rich data about intersections and can encode information such as nominal trajectories and maximum speed of all traffic through a region. Assuming that the map and the location of the ego vehicle are known, an \textit{observable polygon} can be generated for a vehicle's sensor configuration (maximum range, angular resolution, and field of view) without the actual sensor returns. Here we focus on \acp{LIDAR}, but the principles remain the same for other sensor modalities. The shape of the observable polygon is constrained by occlusions caused by objects such as other vehicles, trees, and buildings. With the observable polygon, one can identify free space at the current time. However, the current observable polygon alone can provide little information about long-term risk.

Current free space estimates are insufficient for planning for the future as vehicles can suddenly appear from regions outside of the observable polygon. In order to quantify the risk due to limited sensing in the context of long-term planning, we need to consider vehicles that are potentially hidden in unobserved regions. We leverage the paradigm of the particle filter to perform this prediction. Particles are used to represent the distribution of potential vehicle locations originated from unobserved regions. This approach was selected because of its simplicity and parallelizablility.

Consider a scenario with two lanes shown in Fig. \ref{fig:steps}. We represent the lanes of travel by cubic splines. 
Each cubic spline $c_k$ is parameterized by its position $s$ along the spline:
$$c_k(s) = \begin{bmatrix} x_k(s) \\ y_k(s) \end{bmatrix},s\in [0,\overline{s}_k]$$ $$k\in\{1,2,\ldots,M\}$$
where $[x_k(s)~y_k(s)]^\top$ is the position of a point on $c_k$ at $s$, $M$ is the number of lanes in the scene, and $\overline{s}_k$ is the total length of the spline $c_k$. 

We first extract all possible centroids for all valid vehicle positions in all lanes of travel in the unobserved regions. On each spline $c_k$, we consider $L_k$ disjoint unobserved segments. A set of $N_k$ particles $\left\{(s^{[i]}, v^{[i]})\right\}_{i=1}^{N_k}$ are sampled independently from uniform distributions in these unobserved segments, where $s^{[i]}$ and $v^{[i]}$ are the position and speed of the $i$-th particle. The position $s$ and speed $v$ is distributed as follows: $$s\sim U\left(\bigcup_{j=1}^{L_k} [\underline{s}_j, \overline{s}_j]\right),~v\sim U\left([\underline{v}_k, \overline{v}_k]\right)$$
where $[\cdot, \cdot]$ is a closed set between two real numbers, $U(\cdot)$ is an uniform distribution on a set, $\underline{s}_j$ and $ \overline{s}_j$ is the starting and ending position of an unobserved segment $j$ on spline $c_k$, $\underline{v}_k$ and $ \overline{v}_k$ are the minimum and maximum speed of other vehicles, respectively. Uniform sampling is used because we assume no prior knowledge of $s$ and $v$. A more descriptive distribution can be used if some prior information is available.

Assuming that each particle is traveling with a constant speed, we can then propagate all the particles forward in time for $T_f$ seconds: $$\hat{s}^{[i]} = s^{[i]} + v^{[i]} \cdot T_f$$ where $\hat{s}^{[i]}$ is the position of the $i$-th particle after $T_f$ seconds. This results in a distribution of particles along the centerline of each lane, as shown in Fig. \ref{fig:steps}c. Note that more sophisticated motion models with variable speed can also be used at a cost of more computation time.

To account for the size of vehicles and lateral displacements within the lane, an offset $b^{[i]}$ is sampled from an uniform distribution $U\left([-\overline{b}, \overline{b}]\right)$ and added to each particle in Cartesian space perpendicular to the spline. The uniform distribution is chosen since we assume no prior information for the position of the occluded vehicles in the lateral direction.
$$u_k(s) := \begin{bmatrix}0 & -1 \\ 1 & 0 \end{bmatrix} \cdot \frac{\partial c_k}{\partial s}(s)$$ $$\hat{p}_k^{[i]} = c_k\left(\hat{s}^{[i]}\right) + \frac{b^{[i]}}{\left\|u_k\left(\hat{s}^{[i]}\right)\right\|_2} \cdot u_k\left(\hat{s}^{[i]}\right) $$ where $\|\cdot\|_2$ is the $2$-norm of a vector, $u_k$ is the unnormalized vector perpendicular to $c_k$, and $\overline{b}$ 
is the maximum deviation among all the particles 
from their corresponding centerline. We define the set $\left\{\hat{p}_k^{[i]}\right\}_{i=1}^{N_k}$ to be the distribution of risk over the Cartesian space on lane $k$ after $T_f$ seconds, as shown in Fig. \ref{fig:steps}d.

For observed vehicles, we model them as rectangles along valid lanes. To incorporate them into our proposed formulation, we treat them similarly. For each vehicle, we extract the spline segments within the corresponding rectangle, and apply the aforementioned method as if the segments are in the unobserved regions. This makes our algorithm conservative and performs reasonably well even when the intention and speed of other vehicles are unknown or noisy. If a good tracker is also available, the planner can potentially behaves more aggressively without sacrificing safety. 

The overall distribution of risk $\left\{\hat{p}^{[i]}\right\}_{i=1}^N$ is simply the union of all sets. $$\left\{\hat{p}^{[i]}\right\}_{i=1}^N = \bigcup_{k=1}^M \left\{\hat{p}_k^{[i]}\right\}_{i=1}^{N_k}$$
where $N$ is the total number of particles in the scene on $M$ lanes.


This risk over Cartesian space can be easily integrated with any control or planning algorithm as either the primary cost or in conjunction with other costs as an auxiliary cost function. In addition, it can also be used along with any existing risk assessment method designed for only observed vehicles. In Section \ref{sec:ra-action} we show how we can utilize the risk $\left\{\hat{p}^{[i]}\right\}_{i=1}^N$ as the major cost function of a simple optimization-based planning algorithm.

\subsection{Planning}
\label{sec:ra-action}
In this subsection we demonstrate how the risk described in Section \ref{sec:ra-cartesian} can be used in practice. We integrate it into a optimization-based planning algorithm and show improvements in both safety and ride comfort. The rudimentary planner used here can be replaced by any other cost-based planners as the technique is agnostic to planning approach.

\begin{figure}[t]
  \centering
  \subfloat[Baseline 1]{
    \includegraphics[width=0.499\linewidth,trim={18mm 2mm 2mm 15mm},clip]{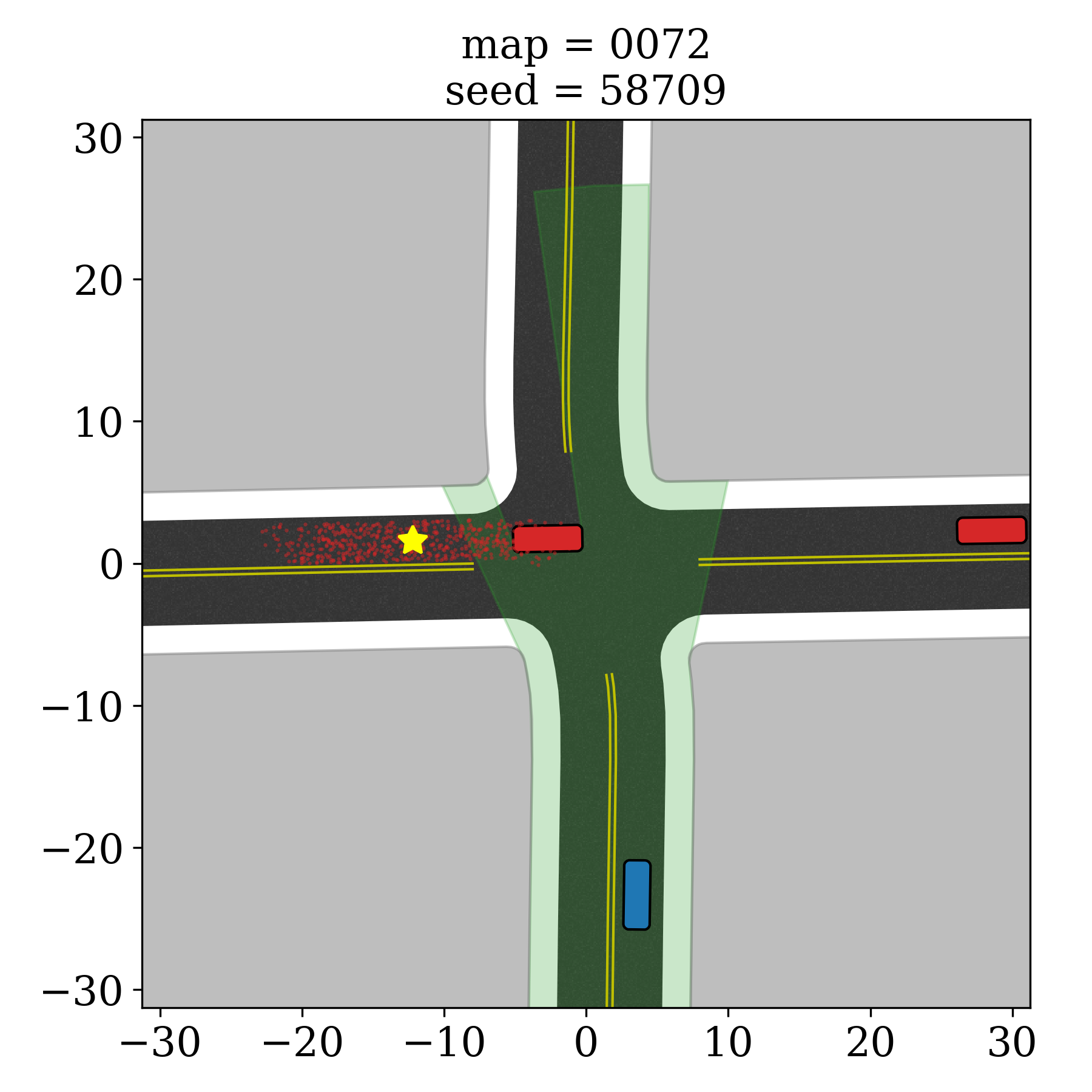}
    \label{fig:ra:baseline}
  }\hspace{-10pt}
  \subfloat[Ours]{
    \includegraphics[width=0.499\linewidth,trim={18mm 2mm 2mm 15mm},clip]{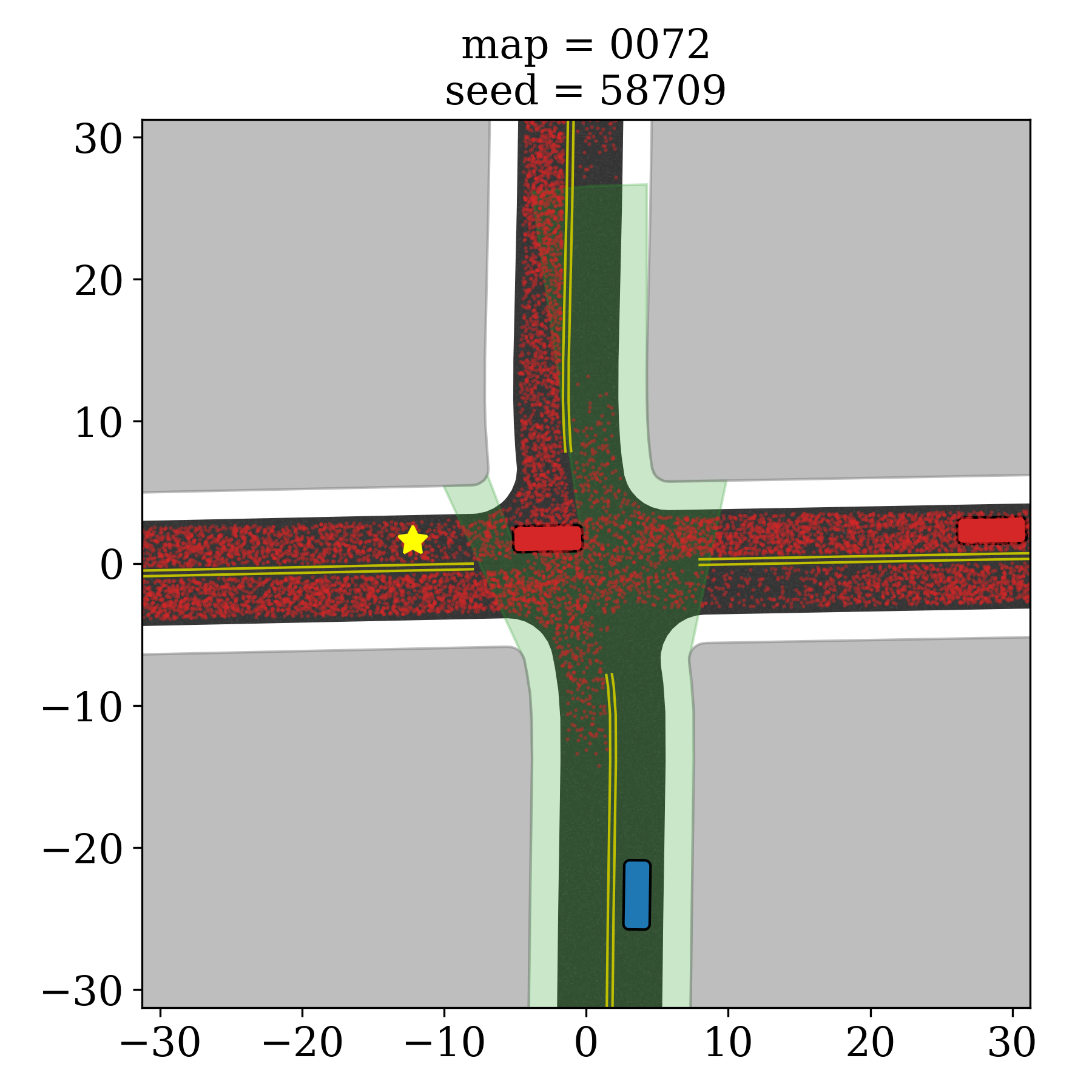}
    \label{fig:ra:proposed}
  } 
  \caption{Comparison between the (a) baseline and (b) proposed method. The baseline method only predicts distribution of risk (red particles) caused by observed vehicles, whereas the proposed method also predicts the risk caused by unobserved regions.}
  \label{fig:ra}
\end{figure}

Assuming that at time $t$ the ego vehicle travels with speed $v_{ego}$ on the intended route $c_{ego}$, which is also a cubic spline parameterized by its position $s_{ego}$ along the spline.
The planner first considers the \textit{safety cost} $J_1(a_{ego})$ associated with an acceleration (or deceleration) $a_{ego}$:
$$J_1(a_{ego}) = \sum_{i = 1}^N f^{[i]}(a_{ego})$$
where $f^{[i]}$ is the potential function of the $i$-th particle which is positive when the particle $\hat{p}^{[i]}$ is within the ego lane and zero otherwise. The function $f^{[i]}$ is defined as follows:
\begin{align*}
    f^{[i]}(a_{ego}) &:=
    \begin{cases}
        \exp\left(-\frac{r^{[i]}(a_{ego})^2}{\sigma^2}\right) &, \mbox{if } d^{[i]}  \leq \overline{b} \\
        0              &, \text{otherwise}
    \end{cases} \\
    \hat{s}_{ego} &:= s_{ego} + v_{ego}\cdot T_f + \frac{1}{2}a_{ego}\cdot T_f^2 \\
    r^{[i]}(a_{ego}) &:= \left\|c_{ego}\left(\hat{s}_{ego}\right) - \hat{p}^{[i]}\right\|_2 \\
    d^{[i]} &:= \inf_{s} \left\|c_{ego}(s) - \hat{p}^{[i]} \right\|_2 
\end{align*} where $\hat{s}_{ego}$ is the future position of the ego vehicle along $c_{ego}$, $r^{[i]}(a_{ego})$ is the distance between particle $\hat{p}^{[i]}$ and $c_{ego}\left(\hat{s}_{ego}\right)$, $d^{[i]}$ is the minimum distance between particle $\hat{p}^{[i]}$ and the ego vehicle's intended route $c_{ego}$, and $\sigma$ is the bandwidth of the repulsive potential field. In practice, particles with $r^{[i]}(a_{ego}) \geq 2\sigma$ are discarded to speed up the calculation.

In addition to the safety cost $J_1(a_{ego})$, a \textit{speed cost} $J_2(a_{ego})$ is also considered to drive the ego vehicle to meet the desired speed $v_{des}$.

$$J_2(a_{ego}) = \left|v_{ego} + a_{ego}\cdot T_f - v_{des}\right|$$ where $|\cdot|$ is the absolute value of a scalar. The optimal acceleration between time $t$ and $t+T_p$ can be found by solving the following optimization problem:
\begin{align*}
\min_{a_{ego}} \hspace*{10mm} & J_1(a_{ego}) + \lambda\cdot J_2(a_{ego}) \hspace*{25mm} \\
    \mbox{ s.t.} \hspace*{1cm} & \underline{v}_{ego} \leq v_{ego} + a_{ego}\cdot T_f \leq \overline{v}_{ego} \\
    & \underline{a}_{ego} \leq a_{ego} \leq \overline{a}_{ego}
\end{align*}
where $T_p$ is the replan time, $\lambda$ is the weight affecting how aggressive the ego vehicle behaves, $\underline{v}_{ego}$ and $\overline{v}_{ego}$ are the minimum and maximum speed of the ego vehicle, and $\underline{a}_{ego}$ and $\overline{a}_{ego}$ are the maximum deceleration and maximum acceleration, respectively. Note that a smaller $\lambda$ favors more conservative behaviors. We find the value of $\lambda$ in Table \ref{tab:param} is a good compromise between safety and efficiency. As shown here, the proposed risk assessment method can be incorporated with any optimization-based planner.

\begin{table}[t]
\caption{Parameters for Simulations}
\label{tab:param}
\begin{center}
    \begin{tabular}{|c|c|} 
        \hline Parameter & Value \\ \hline
        \hline Forecast horizon, $T_f$ & $1.5~s$ \\
        \hline Replan period, $T_p$ & $0.1~s$ \\
        \hline Vehicle length, $l_v$ & $4.88~m$ \\
        \hline Vehicle width, $w_v$ & $1.86~m$ \\
        \hline Number of particles, $N_k~\forall k$ & $\leq 2^{15}$ \\
        \hline Weight, $\lambda$ & $2^{14}\cdot 10^{-6}$ \\
        \hline Bandwidth, $\sigma$ & $0.5 l_v$ \\
        \hline Max. offset, $\overline{b}$ & $0.75 w_v$ \\
        \hline Desired speed, $v_{des}$ & $10~m/s$ \\
        \hline Min. speed, $\underline{v}_k=\underline{v}_{ego}~\forall k$ & $0~m/s$ \\
        \hline Max. speed, $\overline{v}_k=\overline{v}_{ego}~\forall k$ & $12~m/s$ \\
        \hline Min. acceleration, $\underline{a}_{ego}$ & $-8~m/s^2$ \\
        \hline Max. acceleration, $\overline{a}_{ego}$ & $2.5~m/s^2$ \\
        \hline Threshold acceleration, $a_{thresh}$ & $4~m/s$ \\
        \hline
    \end{tabular}
\end{center}
\end{table}

\section{EVALUATION}
\label{sec:evaluation}

To demonstrate how safety and comfort can be improved by our algorithm, we compare it to a baseline approach which only models observed vehicles at intersections. In particular, we focus on scenarios where the ego vehicle tries to make difficult maneuvers such as an unprotected left turn.

\subsection{Simulation}

We simulate various random scenarios with five other vehicles in the scene. Each vehicle travels on a random route at a constant speed ranging from $4$ to $12~m/s$. A valid combination of trajectories is generated by rejection sampling so that there is no collision or overlap among the simulated vehicles.

Here we focus on four-way, un-signaled intersections for compactness and not on T- or Y-junctions, but the proposed approach conceptually generalizes. The layout of intersections can be either synthetic or from real-world map data. For the synthetic layout, the roads are constructed using straight and perpendicular segments. For real-world layouts, we obtain the geometry information from $73$ real-world intersections extracted from \acp{OSM} around Ann Arbor, Michigan. 

To simulate scenarios with heavy occlusion, buildings are added to the map 
with a $2~m$ buffer from the boundary of the driving surface. The ego vehicle starts $15~m$ before the stopline with initial speed $10~m/s$, and tries to perform an unprotected left turn, as shown in Fig. \ref{fig:ra}. More details of the parameters used in the simulator are listed in Table \ref{tab:param}. Note that as length of unobserved segments vary, $N_k$ is calculated dynamically such that the density of particles stays constant at $2^{15}$ particles per $100~m$.

\begin{figure}[t]
  \centering
  \includegraphics[width=0.8\linewidth,trim={0mm 0mm 15mm 0mm},clip]{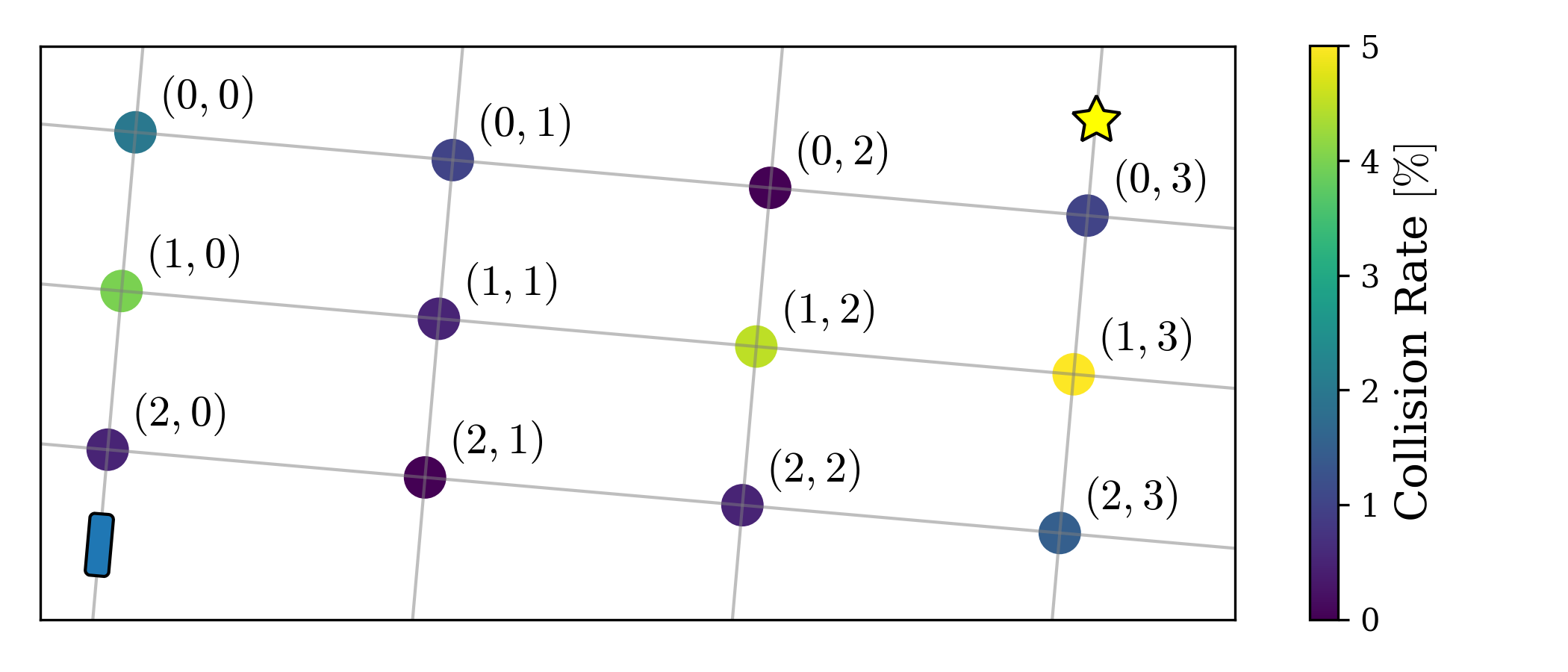}
  \caption{Illustration of collision rates overlaid on a map with $12$ intersection. A high-level planner can plan a route based on the collision rates, taking the route with the lower collision rates: $(2, 0)\rightarrow(2,1)\rightarrow(1,1)\rightarrow(0,1)\rightarrow(0, 2)\rightarrow(0,3)$.}
  \label{fig:collision_map:demo}
\end{figure}

\subsection{Baseline 1}

An occlusion-unaware risk assessment method is used as a baseline for comparison. The baseline method predicts distribution of risk using the exact same method described in Section \ref{sec:ra-cartesian}, but only for the regions intersecting with the observed vehicles. The same planning algorithm described in \ref{sec:ra-action} is used with both the baseline and proposed method throughout all simulations.

\subsection{Baseline 2}

To the best of our knowledge, \cite{orzechowski2018tackling} is the only method that describes both risk assessment and planning in the same article. However, only a few examples were shown in their paper. Here, we recreated the scenarios in their paper and compare the speed and acceleration profiles.

We recreated the scenarios by using the same map, initial speed, desired speed, maximum acceleration and minimum acceleration provided in \cite{orzechowski2018tackling}. The initial position and the speed of the other vehicle in both scenarios are estimated by measuring the plots in \cite{orzechowski2018tackling} using AutoCAD.

\subsection{Metrics}

For baseline 1, we first simulate 2000 random scenarios with the ego vehicle performing an unprotected left turn at each intersection with the baseline method, then simulate the exact same set of experiments with identical trajectories with the proposed method. For each intersection, we calculate its collision rate for both methods as follows:
$$\mbox{Collision Rate} = \frac{\mbox{\# of simulations with collision}}{\mbox{Total \# of simulations}} \times 100\%$$

This meta-collision rate for a given intersection can be used by a high-level planner, which needs to plans a route between two points across a city. Overlaying the collision rates with associated intersections, a high-level planner can avoid dangerous intersections, as show in Fig. \ref{fig:collision_map:demo}.

Speed and acceleration profiles are also calculated to quantify ride discomfort. However, to the best of our knowledge, there is no common computational metric in the literature for ride comfort. Typically, the literature reports thresholds on acceleration and jerk as the metric for ride comfort\cite{hoberock1977survey}. We define the following \textit{discomfort score} to represent a continuous range of discomfort.
$$\mbox{Discomfort Score} = \frac{1}{T}\int_0^T\max\left(0, |a_{ego}(t)| - a_{thresh}\right)dt$$
where $T$ is the duration to reach the goal and $a_{thresh}>0$ is a threshold set to be half of the maximum deceleration.

For baseline 2, speed and acceleration profiles of two scenarios under occlusion are investigated. The first scenario illustrates an ego vehicle traveling at an intersection with no other vehicle present in the scene. The second scenario adds one vehicle coming from the left.

\begin{figure}[t]
    \centering
    \subfloat[\acp{CDF} of collision rates]{
        \includegraphics[width=0.45\linewidth,trim={4mm 0mm 4mm 0mm},clip]{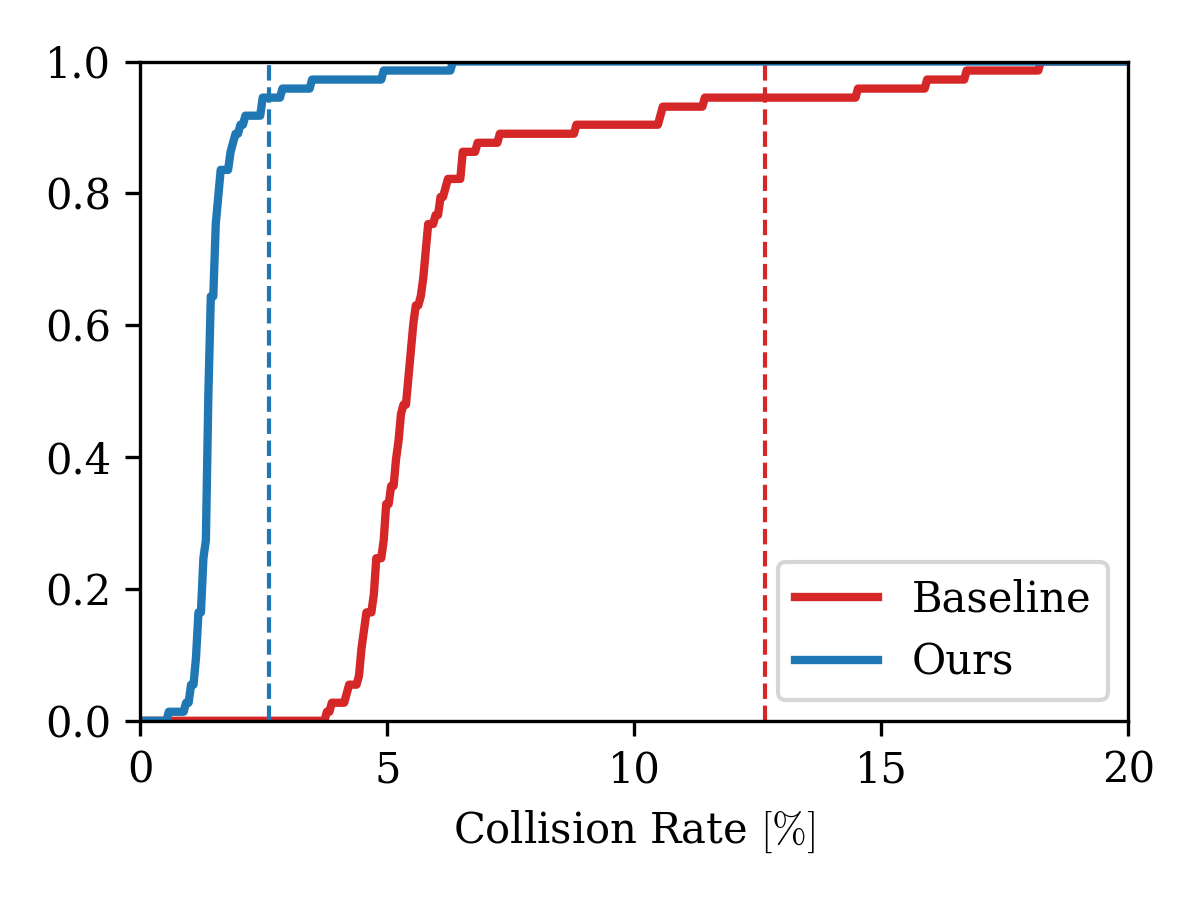}
        \label{fig:rate-cdf}
    }
    \subfloat[\acp{CDF} of discomfort scores]{
        \includegraphics[width=0.45\linewidth,trim={4mm 0mm 4mm 0mm},clip]{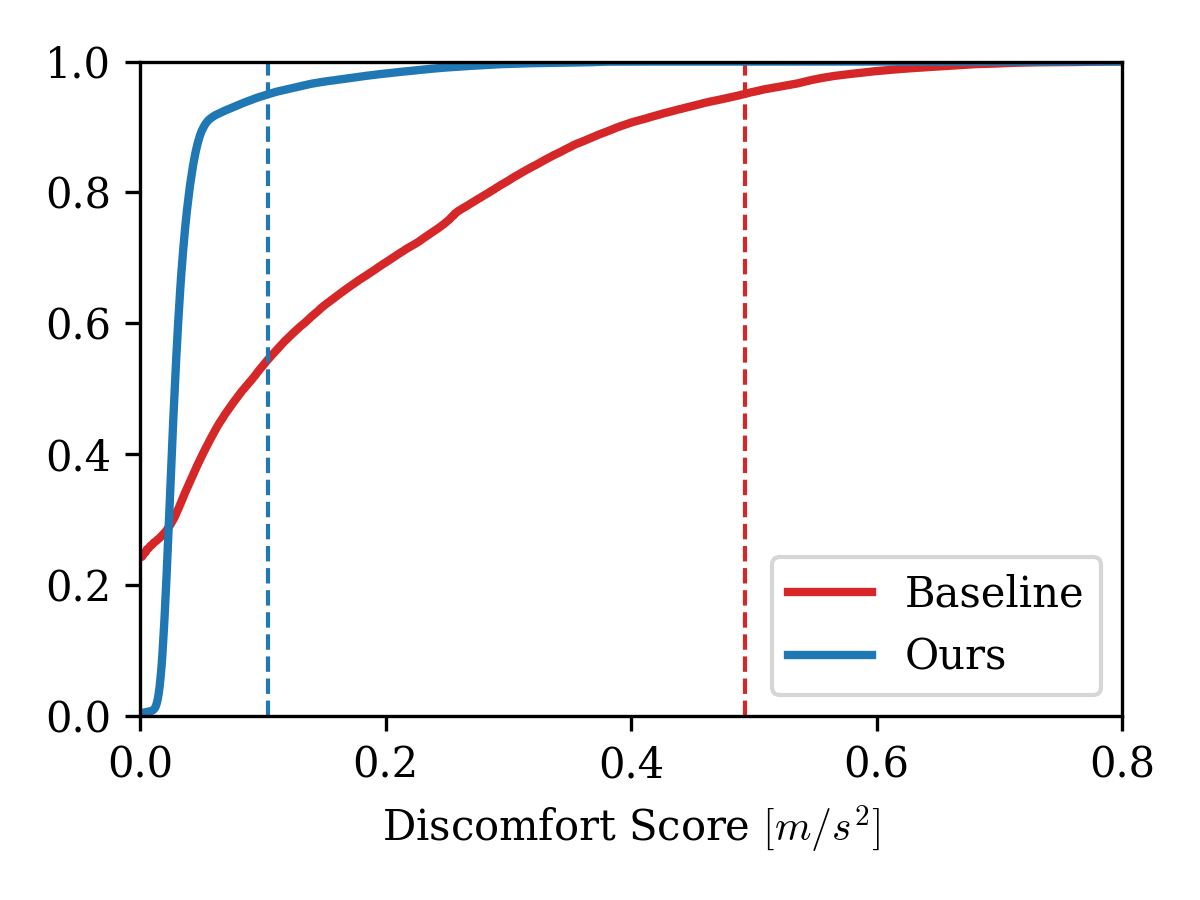}
        \label{fig:discomfort-cdf}
    }
    \caption{\acp{CDF} of (a) collision rates and (b) discomfort scores among all $73$ real-world intersections. Our method outperforms the baseline where the $95$th percentile of the collision rate (vertical dashed lines) decreases from $12.64\%$ to $2.61\%$, and the $95$th percentile of the discomfort decreases from $0.4925$ to $0.1043$.}
\end{figure}

\section{Results}
\label{sec:results}

The distributions of both collision rate and discomfort score are discussed in this section, as shown in Fig. \ref{fig:rate-cdf} and \ref{fig:discomfort-cdf} as \acp{CDF}. In particular, the median and the $95$th percentile are reported. The former describes the nominal behavior, whereas the latter represents the near-worst case.

\subsection{Baseline 1 - Collision Rate}
\label{sec:collision-rate}

We first evaluate the collision rate of both the baseline and proposed method. By utilizing our algorithm, collision rates drop significantly. At the synthetic intersection, the rate decreases by $4.1\times$, from $5.75\%$ to $1.40\%$.
Results at real-world intersections also show similar improvements. Among all the $73$ real-world intersections, the median collision rate decreases by $3.7\times$, from $5.40\%$ to $1.45\%$, and the $95$th percentiles decreases by $4.8\times$, from $12.64\%$ to $2.61\%$, as shown in Fig. \ref{fig:rate-cdf}.

\begin{figure}[t]
  \centering
  \includegraphics[width=0.9\linewidth,trim={5mm 0mm 15mm 0mm},clip]{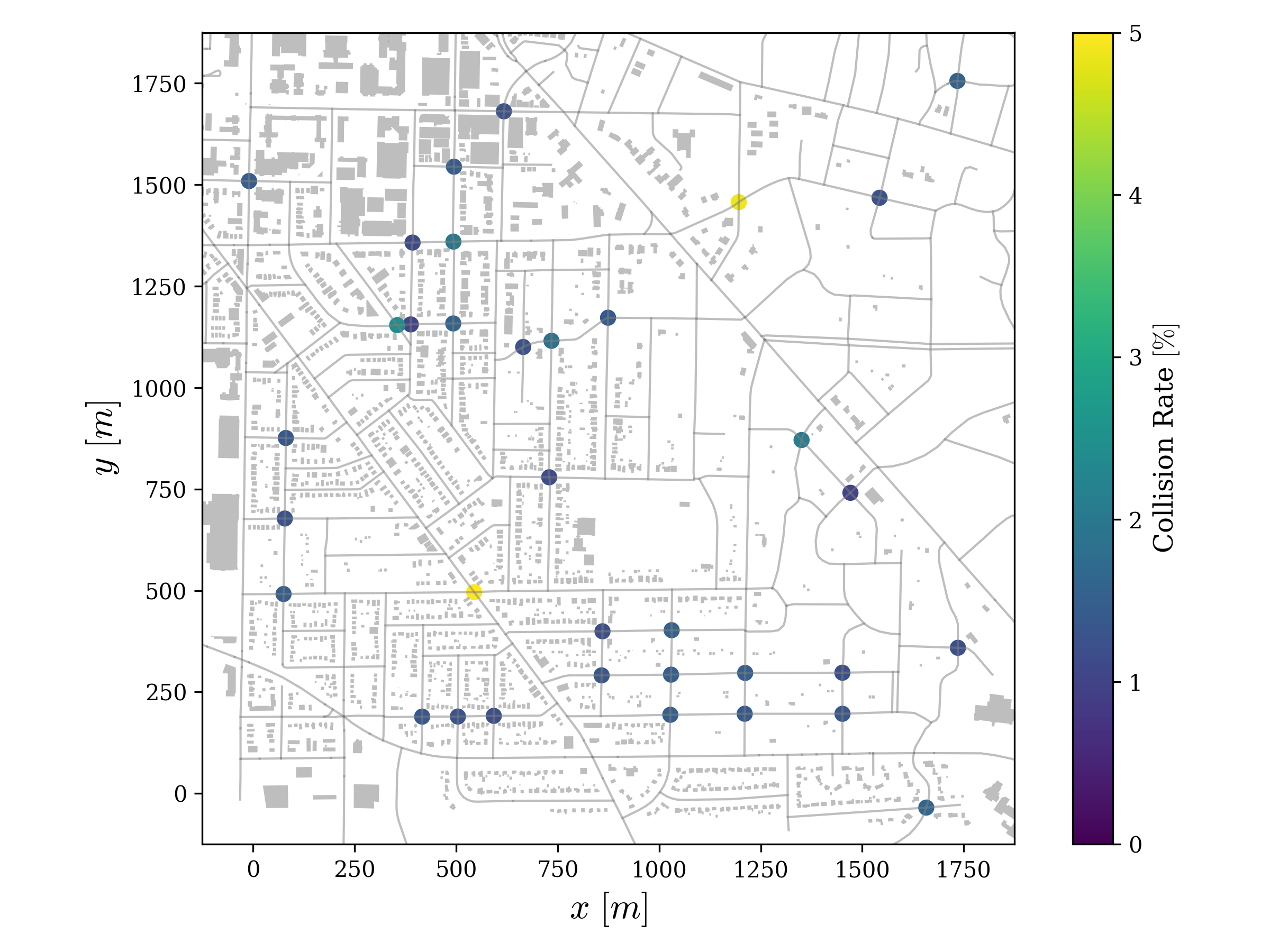}
  \caption{Collision rates of a subset of intersections overlaid on a real-world map. A high-level planner can utilize this information to avoid dangerous intersection such as the two yellow ones.}
  \label{fig:collision_map:aa}
\end{figure}

The distribution of the simulated collision rates of the $73$ real-world intersections is overlaid in Fig. \ref{fig:collision_map:aa}. This can be added as extra information to a high-level planner to plan a route with lower collision rates. This enables a vehicle to reason about safety prior to embarking on a journey. It also enables urban planners and civil engineers to reason about safety of interactions for autonomous vehicles in a systematic and quantitative way.

\subsection{Baseline 1 - Ride Comfort}
\label{sec:ride-comfort}

In addition to safety, which is evaluated as collision rate in Section \ref{sec:collision-rate}, another benefit from our algorithm is more ride comfort. At the synthetic intersection, the median discomfort score is reduced by $2.9\times$, from 0.0795 to 0.0271. The $95$th percentile of the discomfort score decreases by $10\times$, from $0.4687$ to $0.0466$. Similarly, the median score among all real-world intersections is reduced by $3\times$, from $0.0849$ to $0.0284$, and the $95$th percentile of the score decreases by $4.7\times$, from $0.4925$ to $0.1043$.

\begin{figure*}[p!]
  \centering
  \subfloat[Synthetic]{
    \includegraphics[width=0.228\linewidth,trim={0mm 0mm 0mm 13mm},clip]{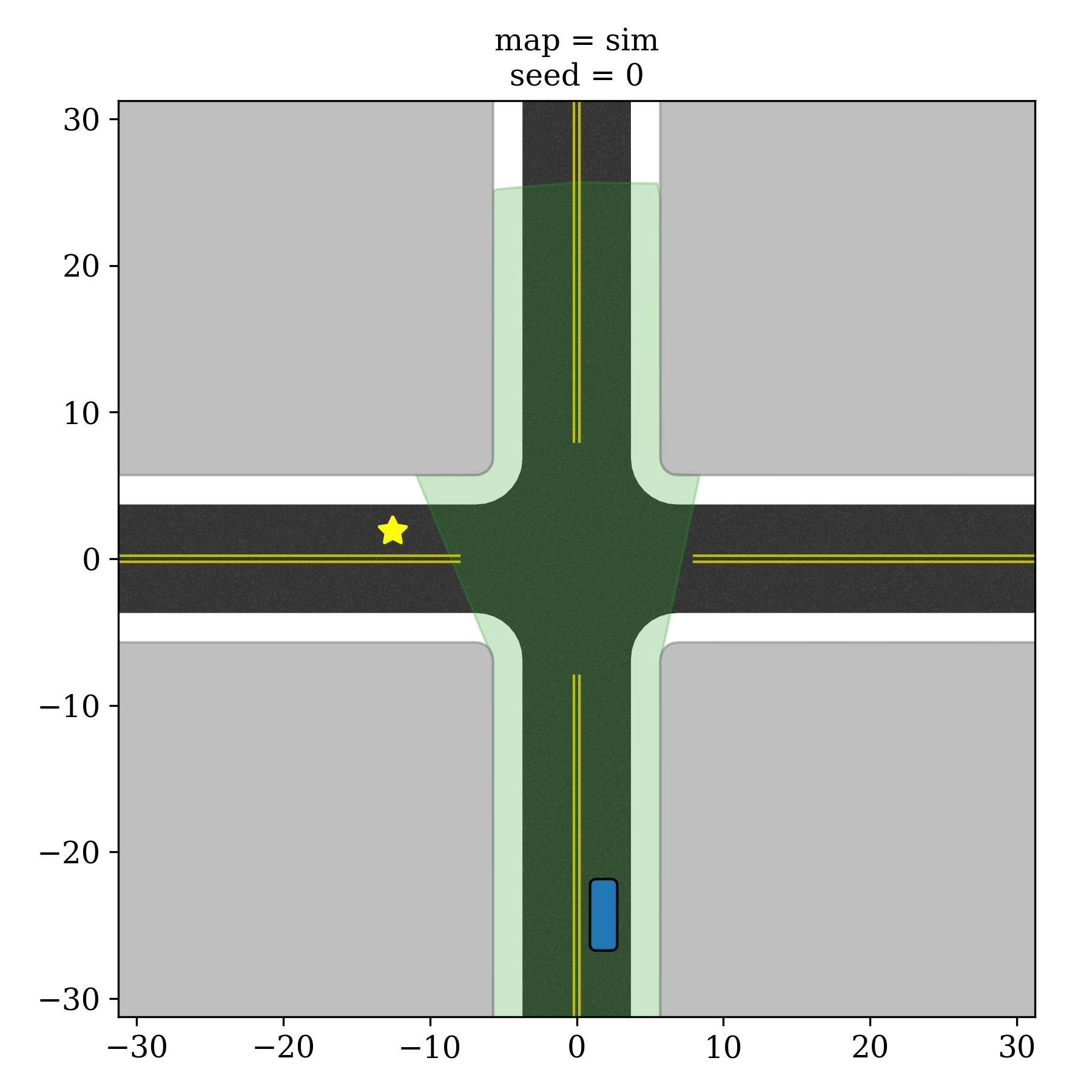}}
  \subfloat[Real-world \#11]{ %
    \includegraphics[width=0.228\linewidth,trim={0mm 0mm 0mm 13mm},clip]{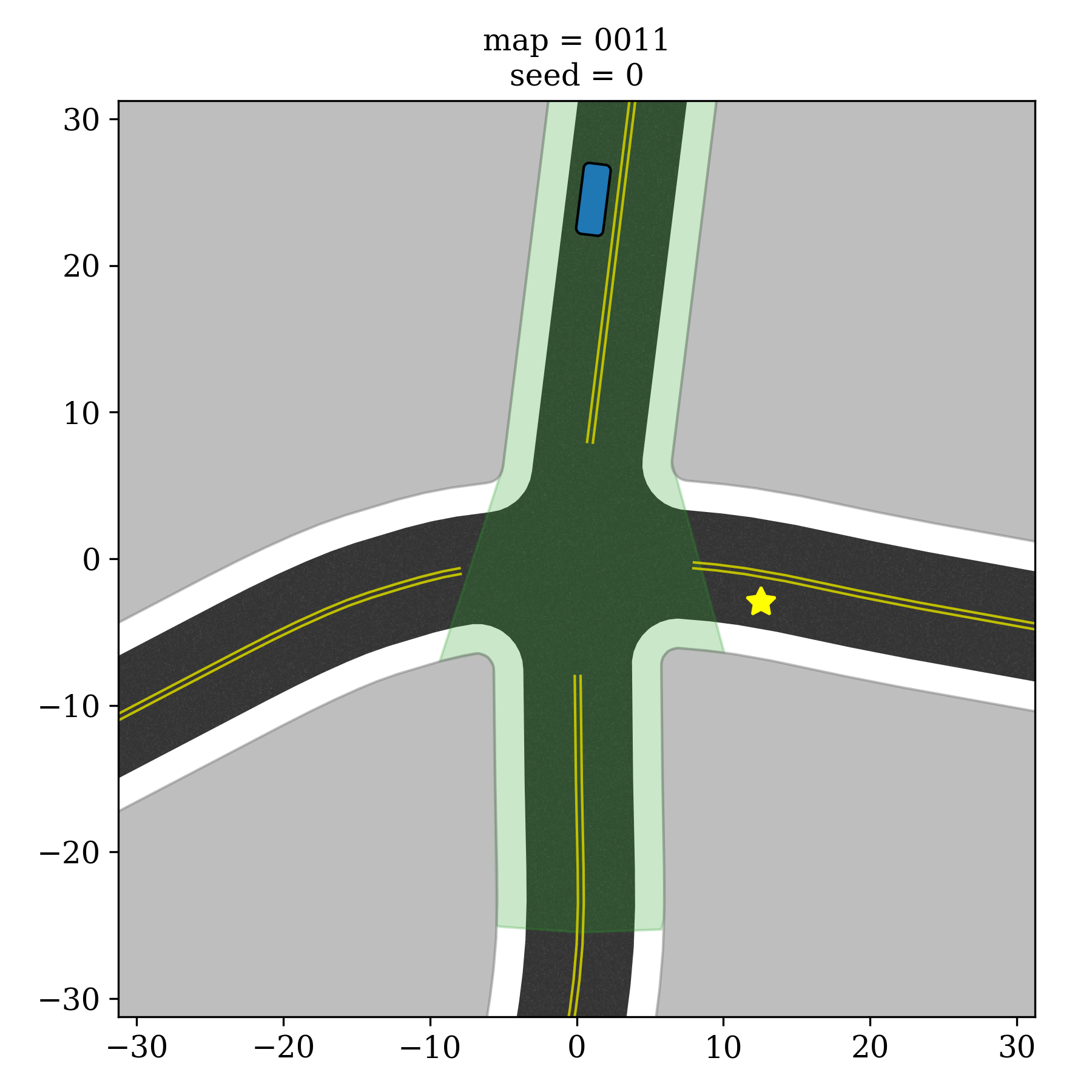}}
  \subfloat[Real-world \#16]{ %
    \includegraphics[width=0.228\linewidth,trim={0mm 0mm 0mm 13mm},clip]{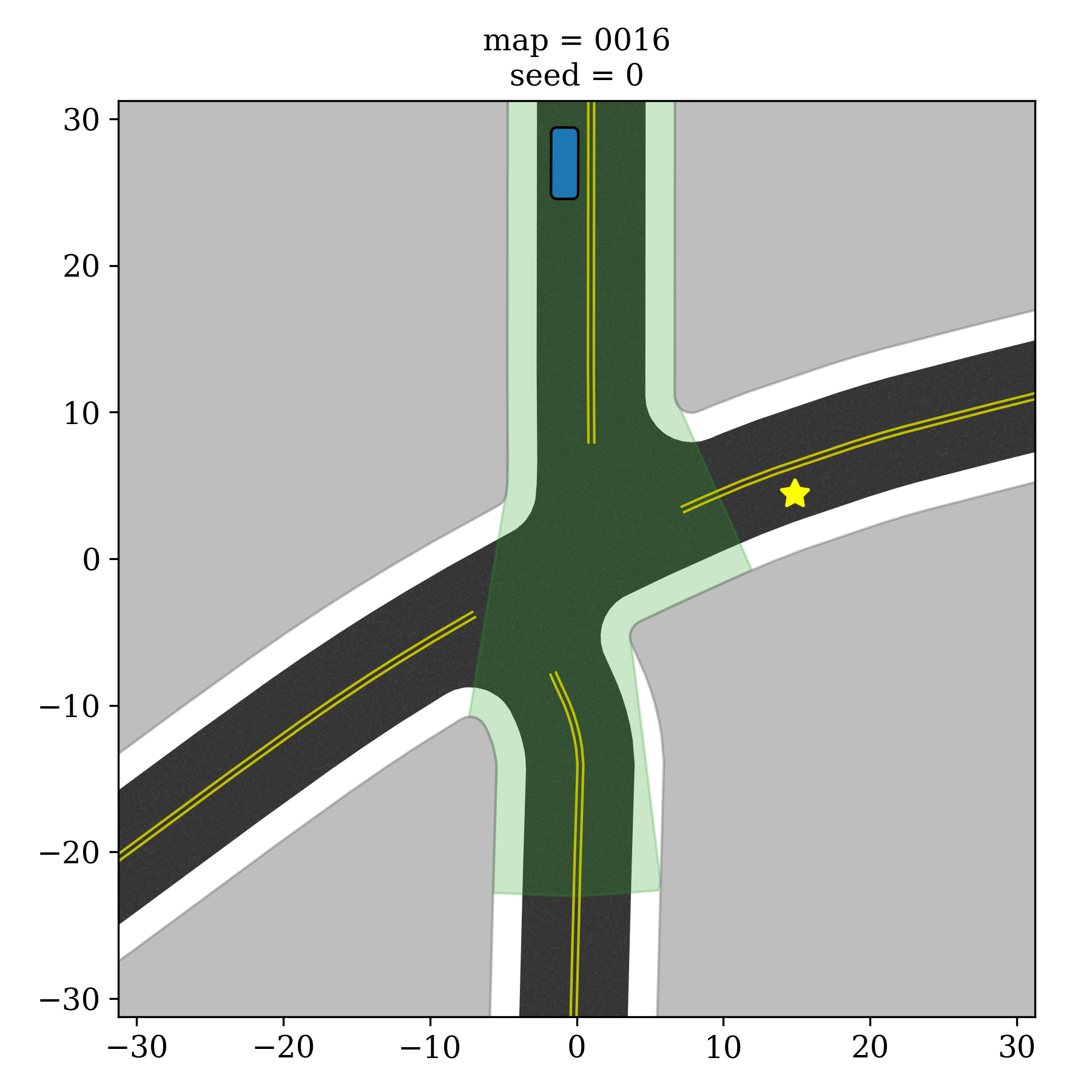}}
  \subfloat[Real-world \#57]{
    \includegraphics[width=0.228\linewidth,trim={0mm 0mm 0mm 13mm},clip]{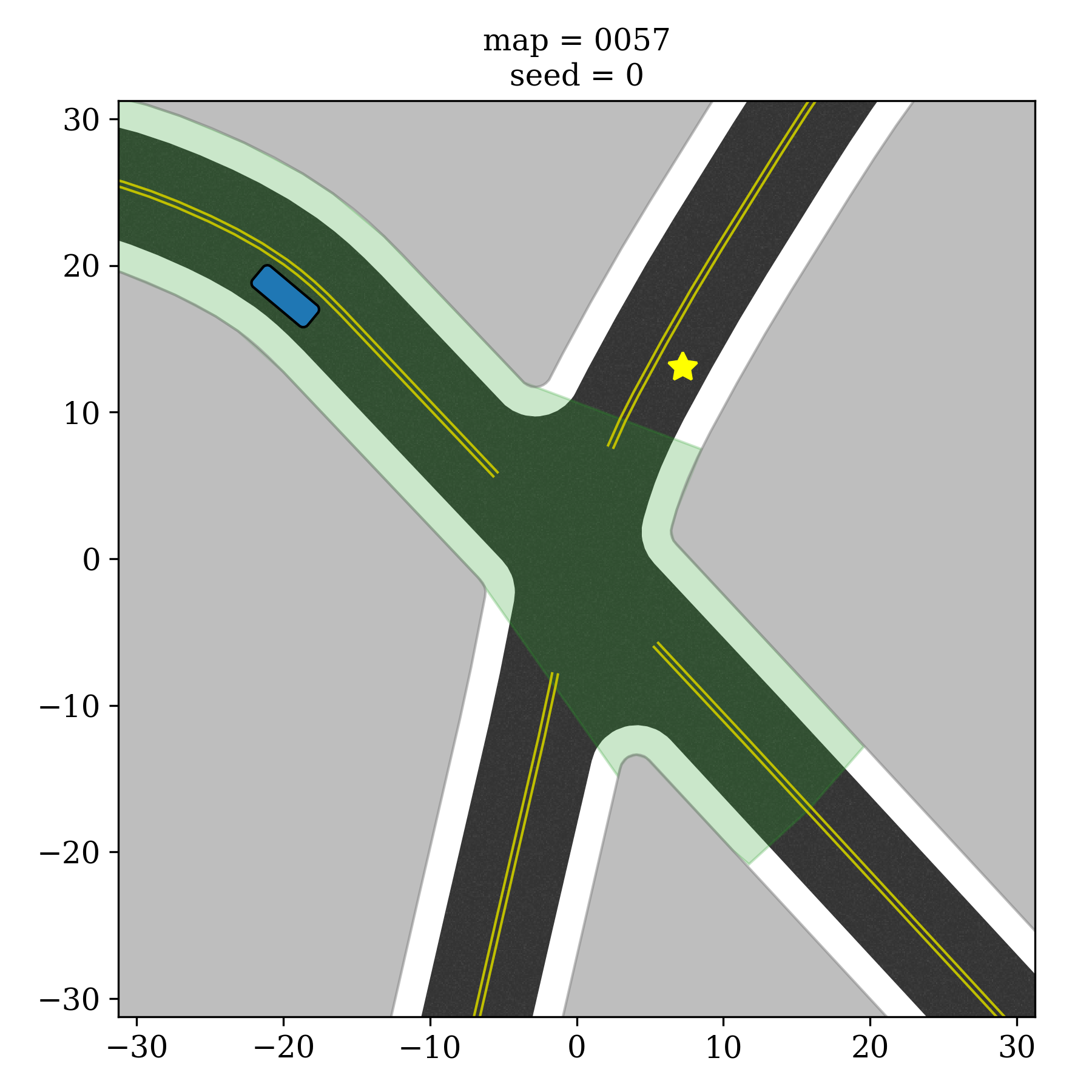}}
    
  \hspace{-20pt}
  \subfloat{
    \includegraphics[height=120mm,trim={0mm 0mm 0mm 4mm},clip]{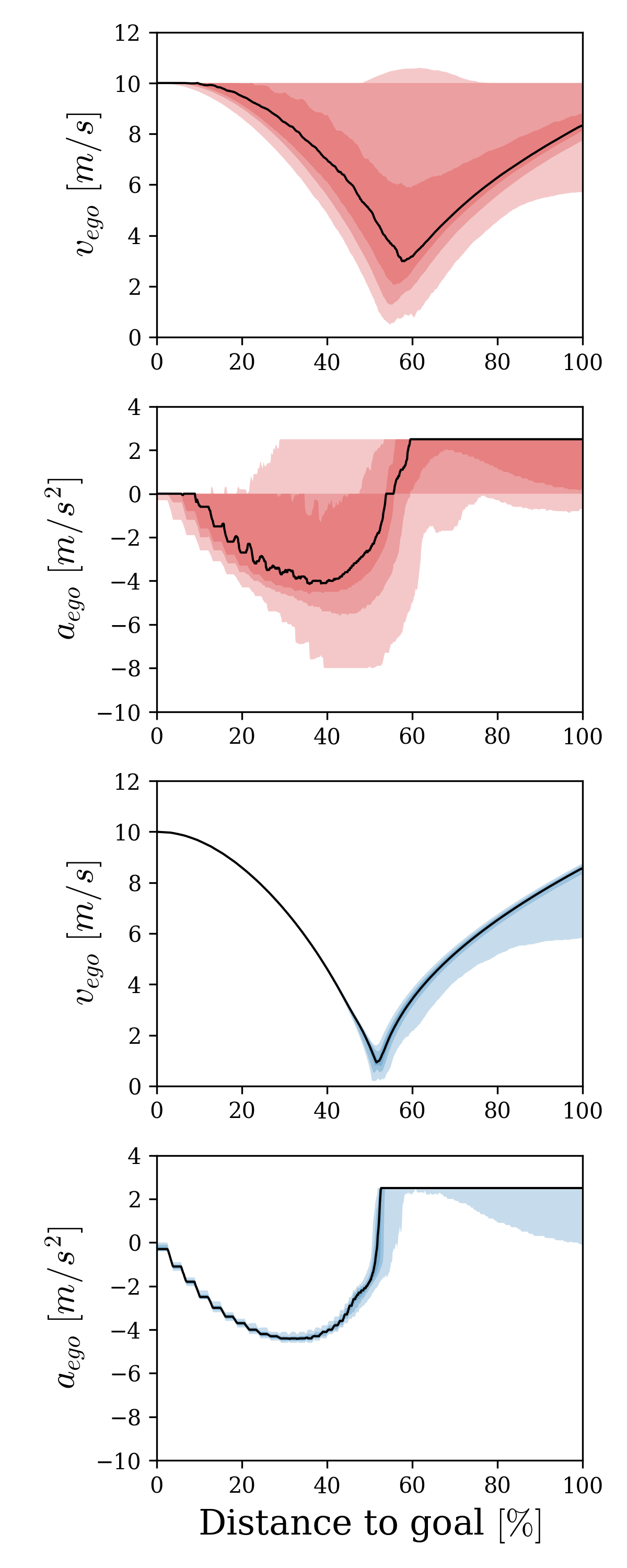}}
  \subfloat{
    \includegraphics[height=120mm,trim={17mm 0mm 0mm 4mm},clip]{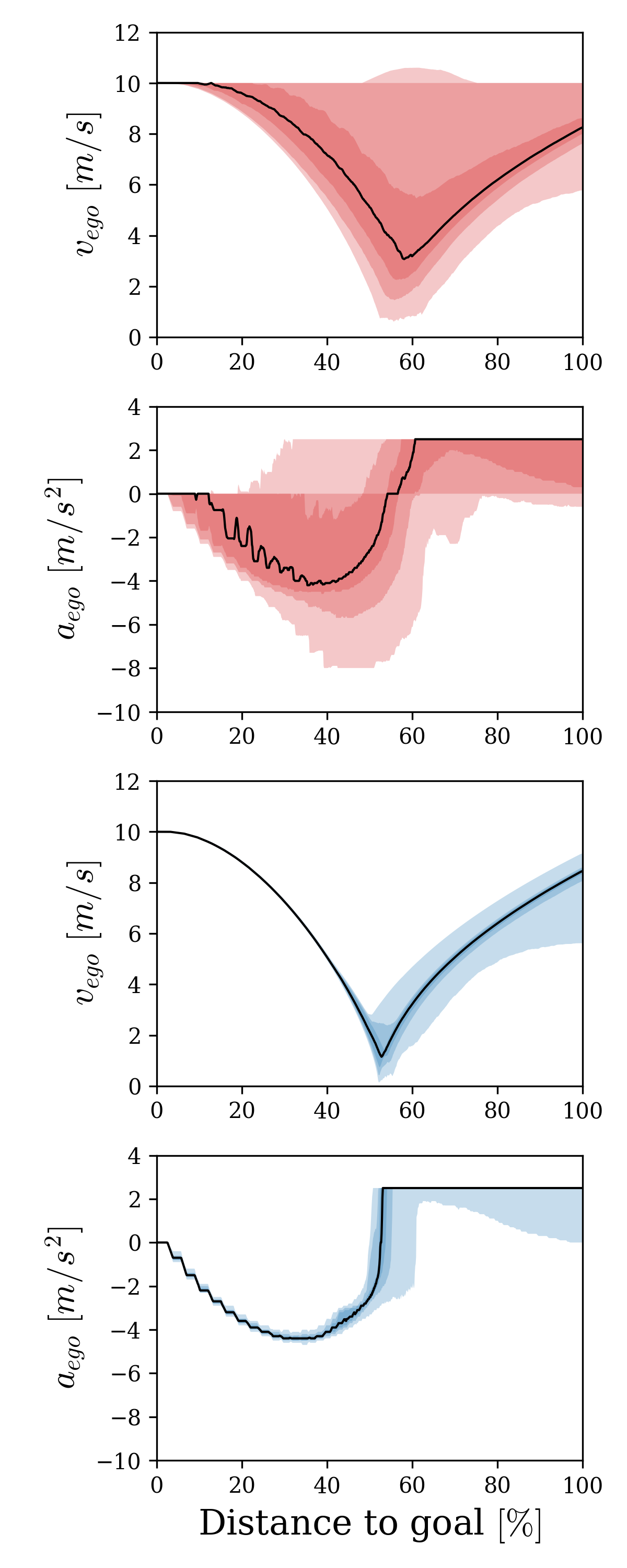}}
  \subfloat{
    \includegraphics[height=120mm,trim={17mm 0mm 0mm 4mm},clip]{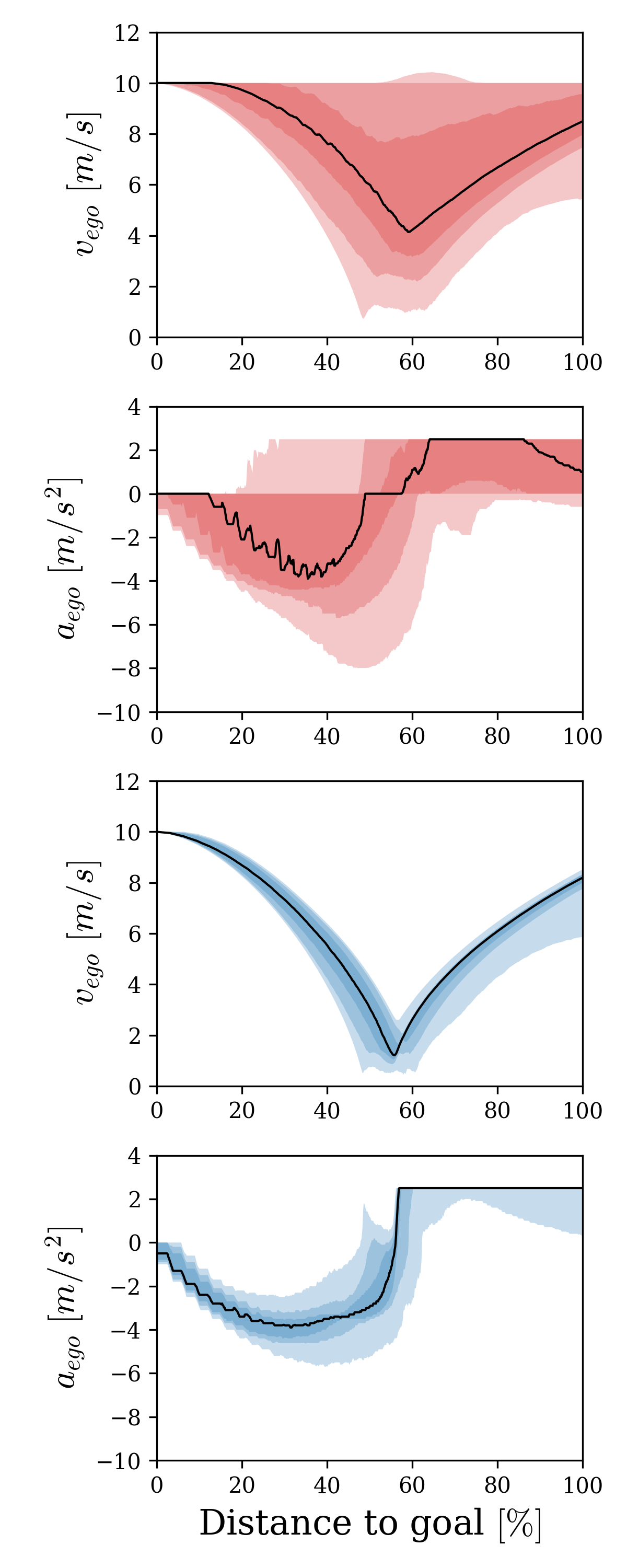}}
  \subfloat{
    \includegraphics[height=120mm,trim={17mm 0mm 0mm 4mm},clip]{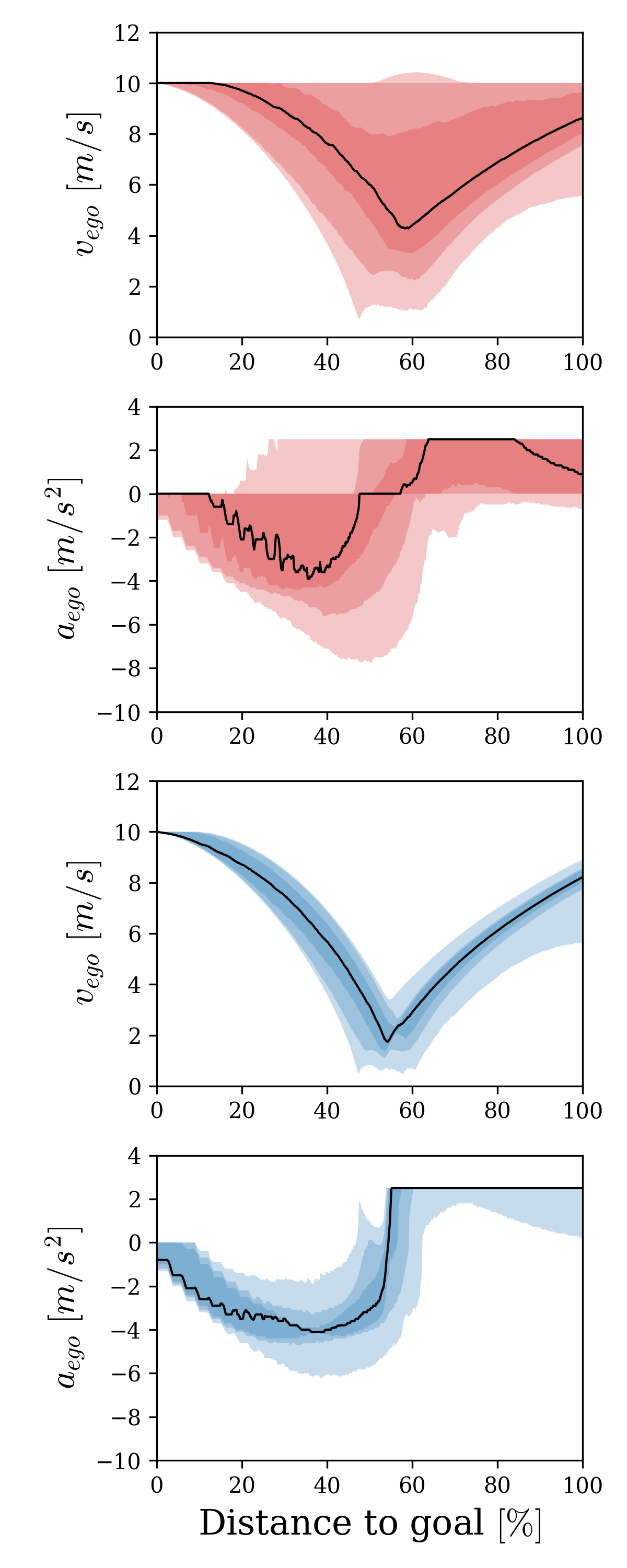}}

  \caption{Speed and acceleration profiles for baseline 1 and the proposed method at various intersections. From top to bottom: map, speed profile of baseline 1, acceleration profile of baseline 1, speed profile of our method, acceleration profile of our method. The medians of profiles are shown in solid black lines, and the percentiles are shown in different shades of colors (from dark to light: $50\pm15\%$, $50\pm30\%$, $50\pm45\%$.) In both synthetic and real-world intersections, the baseline method shows large variations due to abrupt braking. In addition, the deceleration can reach down to $-8~m/s^2$, which can be very uncomfortable. On the other hand, our method predominantly stays above $-4~m/s^2$ and shows smaller variations, which indicates that it performs consistently well across all simulations.}
  \label{fig:profile}
\end{figure*}

An illustration of the synthetic and real-world intersections are shown in Fig. \ref{fig:profile}. In both cases, the baseline method has larger variations in both speed and acceleration, which means that the ego vehicle can abruptly brake only after other vehicles enter the observable polygon. On the other hand, our method naturally introduces a \textit{virtual} stop sign, slowing the ego vehicle down even when there is no other vehicle in the observable polygon, which generates a consistent behavior across all simulations.

\subsection{Baseline 2}

\begin{figure}[t]
    \centering
    \hspace{-20pt}
    \subfloat[Baseline 2 \cite{orzechowski2018tackling}]{
        \includegraphics[width=0.465\linewidth]{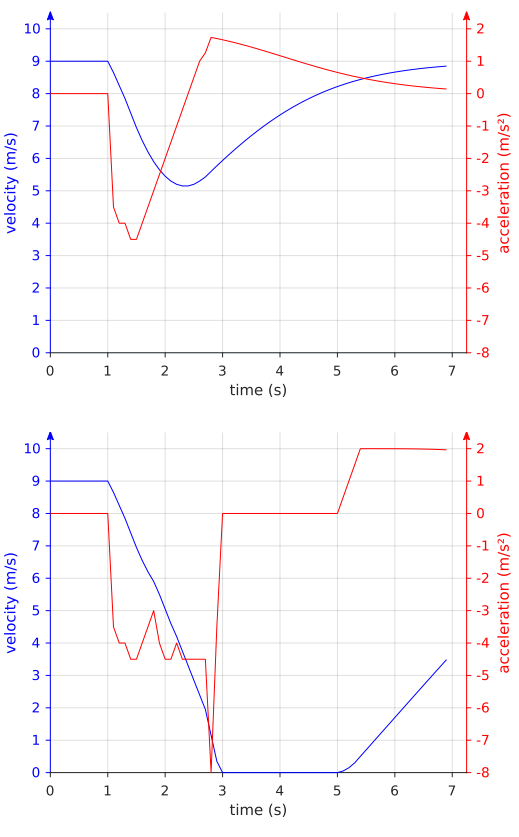}
        \label{fig:baseline:Orzechowski-Meyer}
    } \hspace{-10pt}
    \subfloat[Ours]{
        \includegraphics[width=0.535\linewidth]{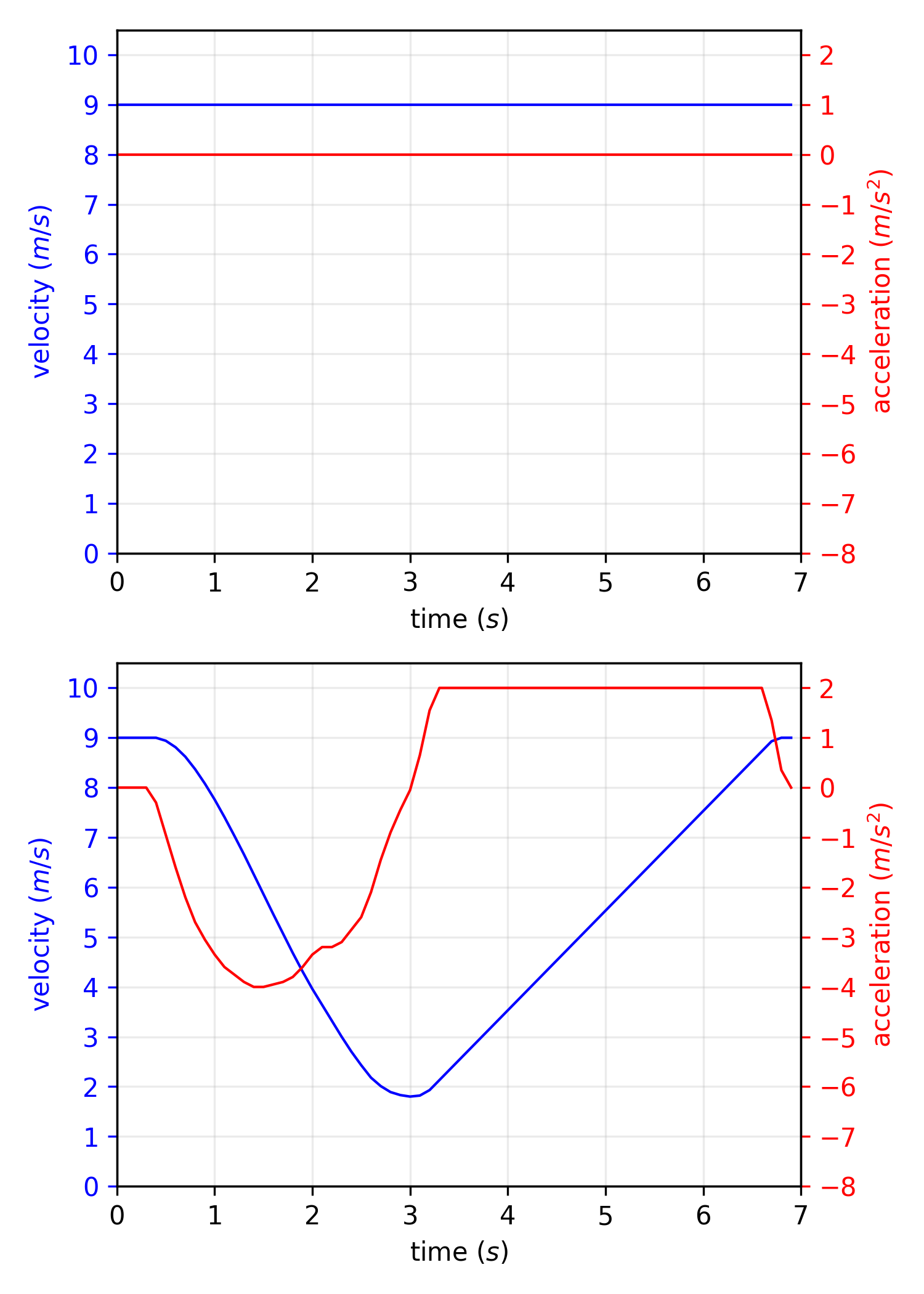}
        \label{fig:baseline:ours}
    }
    \caption{Speed and acceleration profiles of the two scenarios in \cite{orzechowski2018tackling}. Top: with no other vehicle; bottom: with one other vehicle. The left column is adapted from \cite{orzechowski2018tackling}.}
\end{figure}

Both \cite{orzechowski2018tackling} and the proposed algorithm traverse through the intersection safely without collision. But our algorithm performs quantitatively better in terms of ride comfort, as shown in Fig. \ref{fig:baseline:Orzechowski-Meyer} and \ref{fig:baseline:ours}. For the first scenario, our method obtains zero acceleration, whereas the baseline unnecessarily decelerates. The second scenario further highlights the benefit of using a probabilistic risk assessment approach, as opposed to a deterministic one proposed in \cite{orzechowski2018tackling}. The ego vehicles with our algorithm accelerate and decelerate gracefully without any jitter, while the baseline shows large jerk which leads to an uncomfortable ride experience.

\section{CONCLUSIONS}
\label{sec:conclusions}

We propose a probabilistic risk assessment algorithm for autonomous driving under occlusion. We show how it can be integrated with a simple planning algorithm, and compare the proposed algorithm with a baseline method which only performs risk assessment for observed vehicles. We evaluate our algorithm in terms of collision rate and ride comfort with a large number of simulations at one synthetic and $73$ real-world intersections. The results show that the proposed algorithm reduces the collision rate by up to $4.8\times$ and increase comfort by up to $10\times$. Our method shows great potential for improving both safety and comfort for autonomous driving in urban environments.

Future directions include incorporating information from other vehicles for cooperative planning, and handling heterogeneous traffic scenarios with various kinds of vehicles and pedestrians.

\clearpage










\printbibliography

\end{document}